# CenterNet3D: An Anchor Free Object Detector for Point Cloud

Guojun Wang, Jian Wu, Bin Tian, Siyu Teng, Long Chen, *Senior Member, IEEE,* and Dongpu Cao, *Senior Member, IEEE*

*Abstract*—Accurate and fast 3D object detection from point clouds is a key task in autonomous driving. Existing one-stage 3D object detection methods can achieve real-time performance, however, they are dominated by anchor-based detectors which are inefficient and require additional post-processing. In this paper, we eliminate anchors and model an object as a single point—the center point of its bounding box. Based on the center point, we propose an anchor-free CenterNet3D network that performs 3D object detection without anchors. Our CenterNet3D uses keypoint estimation to find center points and directly regresses 3D bounding boxes. However, because inherent sparsity of point clouds, 3D object center points are likely to be in empty space which makes it difficult to estimate accurate boundaries. To solve this issue, we propose an extra corner attention module to enforce the CNN backbone to pay more attention to object boundaries. Besides, considering that one-stage detectors suffer from the discordance between the predicted bounding boxes and corresponding classification confidences, we develop an efficient keypoint-sensitive warping operation to align the confidences to the predicted bounding boxes. Our proposed CenterNet3D is non-maximum suppression free which makes it more efficient and simpler. We evaluate CenterNet3D on the widely used KITTI dataset and more challenging nuScenes dataset. Our method outperforms all state-of-the-art anchor-based one-stage methods and has comparable performance to two-stage methods as well. It has an inference speed of 20 FPS and achieves the best speed and accuracy trade-off. Our source code will be released at *https://github.com/wangguojun2018/CenterNet3d.*

Manuscript received November 18, 2020; revised April 27, 2021 and September 28, 2021; accepted October 1, 2021. This work was supported in part by the Key-Area Research and Development Program of Guangdong Province under Grant 2020B090921003 and Grant 2020B0909050001 and in part by the National Natural Science Foundation of China under Grant 61503380 and Grant 61773381. The Associate Editor for this article was D. F. Wolf. *(Corresponding author: Bin Tian.)*

Guojun Wang and Jian Wu are with the State Key Laboratory of Automotive Simulation and Control, Jilin University, Changchun 130022, China (e-mail: 839977837wgj@gmail.com; wujian@jlu.edu.cn).

Bin Tian is with the State Key Laboratory of Management and Control for Complex Systems, Institute of Automation, Chinese Academy of Sciences, Beijing 100190, China, and also with the School of Artificial Intelligence, University of Chinese Academy of Sciences, Beijing 100190, China (e-mail: bin.tian@ia.ac.cn).

Siyu Teng is with the Department of Computer Science, Hong Kong Baptist University, Hong Kong (e-mail: tengsyslash@gmail.com).

Long Chen is with the School of Data and Computer Science, Sun Yat-sen University, Guangzhou, Guangdong 510275, China (e-mail: chenl46@mail.sysu.edu.cn).

Dongpu Cao is with the Department of Mechanical and Mechatronics Engineering, University of Waterloo, Waterloo, ON N2L 3G1, Canada (e-mail: dongpu.cao@uwaterloo.ca).

Digital Object Identifier 10.1109/TITS.2021.3118698

## I. INTRODUCTION

THE 3D object detection is the task to recognize and locate objects in 3D scenes. It serves as a fundamental task for 3D scene understanding with wide applications in autonomous driving. Recent approaches utilize various types of data, including monocular images [1]–[4], stereo images [5], [7] and point cloud from Radar [8]–[10] and LiDAR [11], [12]. Mono3D [1] uses context, semantics, and hand-engineered shape features to generate proposals, and the proposals are further scored and refined by a Fast-RCNN. Mousavian *et al.* [2] proposed to regress 3D object properties using a convolutional neural network and then combined these estimates with geometric constraints to produce 3D bounding boxes. Chen *et al.* [5] proposed a 3D object proposal method that uses 3D information estimated from a stereo camera pair to generate high-quality 3D object proposals. Palffy *et al.* [8] proposed a novel Radar-based, single-frame, multi-class object detection method for moving road users (pedestrian, cyclist, car), which utilizes low-level Radar point cloud data.

Compared with 2D images, the point cloud from LiDAR has accurate depth information. In addition, compared with Radar, LiDAR has higher accuracy and resolution which can produce denser point cloud and high-precision 3D images. Thus, LiDAR has become an indispensable sensor for autonomous driving. Due to the operation mechanism of LiDAR, point clouds are sparse and unordered which making them impossible to directly use convolution neural networks (CNNs) to parse them. Therefore, how to convert and utilize raw point cloud data has become the primary problem in 3D detection tasks.

In order to handle these problems, many existing methods convert point clouds from sparse points to compact representations with 2D/3D voxelization, and a PointNet feature extractor is applied in each voxel. Then the learned voxel features are converted to 2D/3D pseudo images, these methods achieve both good accuracy and inference speed. The VoxelNet [11], along with its successors, SECOND [12], PointPillar [13] and SA-SSD [15] densely place anchor boxes over the final feature maps and classify them directly. And the offsets from the anchors are predicted to generate bounding boxes.

But the use of anchor boxes has three drawbacks. First, a large number of anchor boxes are needed to place to overlap different aspect ratios and scales which introduces the extra burden. Thus, they require many hyper-parameters and design choices, such as anchor ranges, anchor sizes, and orientations. Second, the IOU matching threshold must be carefully determined to obtain the appropriate positive and negative samples, and the performance of the networks is very sensitive to it. Third, Non-Maximum Suppression (NMS) is necessary for anchor-based methods to suppress the overlapped bounding boxes. So it also introduces extra computational cost which is not conducive to model deployment in practical applications.

Recently, academic attention has been geared toward anchor-free detectors due to the emergence of FPN [16] and Focal Loss [17], such as [18]–[21] and [22]. These anchor-free detectors directly find objects without pre-defined anchors in two different ways. One way is to choose several pre-defined keypoints as positive samples, this type of anchor-free detector is keypoint based method [20]–[22]. Another way is to use the center region of objects to define positive samples, this kind of anchor-free detector is the anchor point method [18], [19]. CenterNet is the first non-maximum suppression free 2D detector based on key points, it is natural to expand its ideas for 3D point cloud, just like AFDet [23].

Also inspired by CenterNet [22], we propose a CenterNet3D Network that performs 3D object detection to generate oriented 3D bounding boxes without the predefined anchors in 3D point cloud. Our method detects objects in LiDAR point clouds by representing them as center points of bounding boxes. The properties of bounding boxes such as size and orientation are then regressed directly from feature maps at the center locations.

Since LiDARs can only capture surfaces of objects, 3D object centers are likely to be in empty space. This inherent sparsity of point clouds makes it difficult to learn effective scene context only from the center point. Besides, the center point may be far from the boundary of an object, so it is difficult to estimate accurate boundaries. To solve this problem, we propose an extra corner attention module to predict the corner points. The corner attention module can enforce the CNN backbone to pay more attention to object boundaries which is helpful to learn more effective corner heatmap and regress more accurate bounding boxes.

In addition, one-stage detectors often suffer from the discordance between the predicted bounding boxes and corresponding confidences. To alleviate the misalignment between the localization accuracy and classification confidence without having an additional network stage, we design an efficient keypoint-sensitive warping operation, namely KSWarp, for post-processing the confidence. The proposed KSWarp can make full use of the predicted corner heatmap and the center heatmap to generate a confidence heatmap that is more consistent with localization accuracy.

Because we only have one positive sample per object, inference is a single network forward-pass, without non-maximum suppression for post-processing. Peaks in confidence heatmap correspond to object centers. The main contributions of our proposed method can be summarized as follows:

1. We propose a novel 3D object detection head CenterNet3D head for point cloud, each object is represented by a center point of its bounding box. Other properties, such as object size and orientation are then regressed directly at each center location. Inference is a single network forward-pass, without non-maximum suppression for post-processing.

2. We propose a corner attention module to enforce the CNN backbone to learn discriminative corner features, which makes the network aware of object size and structure information.

3. We develop an efficient feature map warping method to make the classification confidence more consistent with the localization accuracy, improving the detection performance at negligible cost.

4. Based on CenterNet3D head, we architect the CenterNet3D network for LiDAR point clouds and this is a keypoint based anchor-free 3D detection method which achieves competitive accuracy compared with anchor-based detectors on the KITTI and nuScenes datasets.

The remainder of this paper is organized as follows. Section II describes a review of related research. Section III introduces the methodology for our CenterNet3D detection method. Section IV presents the results of comprehensive experiment evaluations. Section V summarizes the major contributions of this research and future work.

## II. RELATED WORK

Currently, there are four types of point cloud representations as input for 3D detectors. 1) point-based representation [24]–[27]. The raw point cloud is directly processed, and bounding boxes are predicted based on each point. 2) voxel-based representation [8], [12], [13], [28], [29]. The raw point clouds are converted to compact representations with voxelization. 3) Mixture of representations [30], [31]. In these methods, both points and voxels are used as inputs, and their features are fused at different stages of the networks for bounding box prediction. Different methods may consume different types of point cloud representation, in this paper, we adopt voxel-based representations to obtain a tradeoff between efficiency and accuracy. Below, we will briefly review one-stage and two-stage 3D object detection methods, and then we emphasize possibly related anchor-free 3D object detection.

### A. One-Stage 3D Detection

One-stage 3D detectors are proposed to enhance the computational efficiency by processing the input data once in a fully convolutional network. First, a contiguous and regular feature representation is constructed by 2D/3D voxelization. Then a 2D/3D CNN backbone is applied to extract features to obtain 3D bounding boxes. Zhou *et al.* proposed to extracted voxel feature by a simplified PointNet [47]. On the observation of sparsity in voxels, [12] proposes a sparse convolution algorithm and integrates it into the original framework of [8] to accelerate the calculation of convolutional layers. To improve computational efficiency further, PointPillars [13] further simplifies SECOND by implementing voxelization only in the Birds Eye View (BEV) plane. To recover spatial and structure information, [15] proposes a detachable auxiliary network to learn structure information and exhibits better localization performance without extra cost. Our proposed method is also built on top of general architecture, similar to [12] and [15], to reduce the complexity of computation.

### B. Two-Stage 3D Detection

Unlike one-stage approaches that directly produce 3D bounding boxes, two-stage approaches firstly generate plausible candidate proposals, then more accurate bounding boxes are obtained by re-using the point cloud within these candidate proposals in the second stage. Some image-driven methods [42], [41] have been proposed to lift a set of 3D regions of interest (ROI) from the image and then apply a PointNet to extract ROI features by gathering the interior points with transformed canonical coordinates. Shi *et al.* [24] proposed to generate 3D ROIs from raw point clouds by using a PointNet++ backbone. Its variant work [30] generated ROIs by an efficient 3D CNN. Shi *et al.* [43] enriched the ROI features by performing intra-object part-aware analysis and demonstrated the effectiveness of reducing the ambiguity of the bounding boxes. To reduce the computational burden, [26] seeded each point with a new spherical anchor to achieve high recall with less computation. The inherent complex design of these two-stage methods makes them not applicable for autonomous driving.

### C. Anchor-Free 3D Detection

Most of the one-stage and two-stage approaches above have relied heavily on the design of pre-defined anchors or pre-defined object sizes which require extra computational burden and hyper-parameters. To our knowledge, there are few anchor-free 3D detectors for LiDAR-based point clouds. SGPN [44] used a single network to predict point grouping proposals and corresponding semantic classes. Then a similarity matrix was learned to group points together. This method is not scalable since the size of similarity matrix grows quadratically with the number of points. 3D-BoNet [45] directly regressed 3D bounding boxes for all instances in a point cloud, while simultaneously predicting a point-level mask for each instance. VoteNet [46] used PointNet++ [47] to generate seed points and independently generated votes through a shared voting module. Then the clustered points were refined to obtain box proposals. Though with some anchor-free flavor, VoteNet is not strictly anchor-free because it uses anchor boxes for the size regression, similar to Point-RCNN [24]. HotSpotNet [36] used the voxels located in the center region of bounding boxes as positive samples and then directly regressed bounding boxes. It eliminates the anchors but still requires additional non-maximum suppression post-processing. AFDet was the first real anchor-free 3D object detection method, it also used center points to represent objects.

## III. CenterNet3D Network

Our CenterNet3D represents LiDAR objects as center points of their bounding boxes. Each center point likelihood is individually predicted on each pixel of CNN feature map. For the voxelization representation, a voxel is a positive sample if the center point of a bounding box resides in. Each voxel can be projected to a neuron on the feature map based on its location. During training, neurons on the feature map are assigned as centers and non-centers for each object category which are trained by a binary classifier. In inference, a neuron on a feature map is considered as a center point if it gives a local peak in a predicted confidence heatmap.

The proposed CenterNet3D consists of a 3D feature extractor, a 2D feature extractor, and CenterNet3D head. CenterNet3D head has three modules for center classification, box regression, and corner classification. Our CenterNet3D head can be stacked on top of any voxel-based detectors for point clouds. The whole architecture of our proposed CenterNet3D is shown in Fig. 1. The input LiDAR point clouds are voxelized into regular grids $I \in R^{L \times W \times H}$ of size $L$, $W$, $H$ along the $X$, $Y$, $Z$ axes respectively for a frame of point cloud. The regular grids $I$ pass through the 3D CNN to generate the 3D feature maps. Then the 3D feature maps are transformed into 2D feature maps by collapsing in $Z$ axis and passed into 2D feature extractor. Finally, the three subnets will guide the supervision and generate the predicted 3D bounding boxes. The center points and corner points assignment happens at the last feature maps of the 2D feature extractor. The details of network architecture and the three subnets for supervision will be described below.

### A. CenterNet3D Head

Our CenterNet3D head consists of three modules: 1) a center classification module that predicts the probability of center points for an object category. 2) a box regression module that generates an eight-channel feature map to regress the properties of bounding boxes: offset regression that regresses the discretization error of center 2D locations caused by down-sampling stride; z-coord regression that predicts the center location in $Z$ axis; size regression that regresses the 3D size of objects; direction regression that regresses the rotation angle around $Z$ axis. 3) a corner classification module that predicts the probability of corner points for an object category. For box regression, there are two alternative methods to implement. The direct way is to predict all the box properties with one common module. Alternatively, we can predict different properties with different modules to learn the specific appearance characteristics of an object instance. We term the two variants of our network according to the way to regress box properties that regressing all box properties with one module or different modules as CenterNet3DMerge and CenterNet3DSplit.

*1) Center Classification:* The center classification outputs center heatmaps with several convolutional layers and each center heatmap corresponds to one category. let $\hat{Y} \in [0, 1]^{\frac{L}{R} \times \frac{W}{R} \times C}$ be the center heatmaps, where $R$ is the downsampling stride and $C$ is the number of center point types which equals the number of classes. The output prediction $\hat{Y}$ is downsampled by a factor $R$. A prediction $\hat{Y}_{x,y,c} = 1$ corresponds to a detected center point, while $\hat{Y}_{x,y,c} = 0$ is background. We generate ground truth center heatmaps $Y \in [0, 1]^{\frac{L}{R} \times \frac{W}{R} \times C}$ following Law and Deng [20]. For each ground truth center point $p \in R^2$ of class $c$, we compute a low-resolution equivalent $\tilde{p} = \left\lfloor \frac{p}{R} \right\rfloor$. We then splat all ground truth center points onto the heatmaps $Y \in [0, 1]^{\frac{L}{R} \times \frac{W}{R} \times C}$ using a gaussian kernel $Y_{xyc} = \exp(-\frac{(x-\tilde{p}_x)^2+(y-\tilde{p}_y)^2}{2\sigma_p^2})$, where $\sigma_p$ is an object size-adaptive standard deviation which is 1/6 gaussian radius in our work. We determine the gaussian radius by the size of an object by ensuring that one point within the radius would generate a bounding box with at least $t$ IoU with the ground-truth annotation. Because those false center predictions that are close to their respective ground truth locations can still produce bounding boxes that sufficiently overlap the ground-truth box, a penalty-reduced pixelwise logistic regression with focal loss is used as a training objective [17].

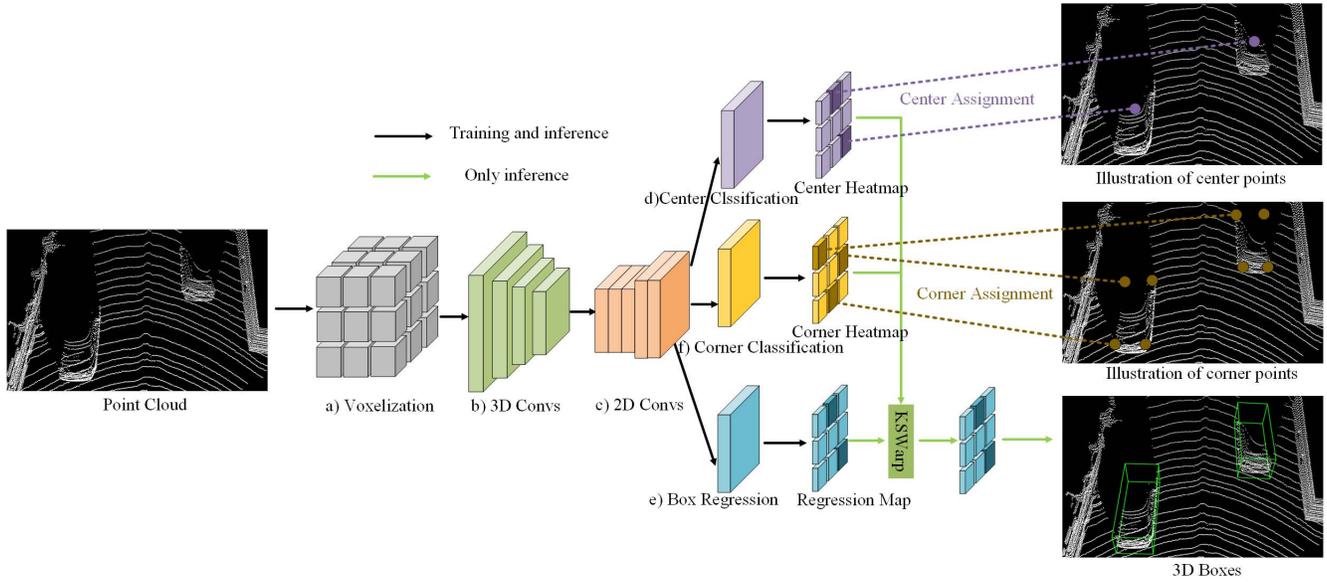

Fig. 1. Outline of CenterNet3D. The point cloud is (a) voxelization (b) 3D Convs including sparse convolution and submanifold convolution proposed in [12]. c) 2D Convs including transposed convolution and deformable convolution to produce 2D feature maps. These feature maps pass into three modules to perform d) Center Classification e) 3D Bounding Box regression (f) Corner Classification to train the network. During the inference the predicted center heatmap and corner heatmap are processed by KSWarp operation to obtain a confidence heatmap that is more consistent with localization accuracy. The final bounding boxes are obtained by processing the confidence heatmap through max pooling, without the need of IOU-based Non-Maximum Suppression (NMS).

$$
\begin{aligned}
&L_{cls} \\
&= \frac{-1}{N} \sum_{xyc}
\begin{cases}
(1-\hat{Y}_{xyc})^\alpha \log\left(\hat{Y}_{xyc}\right) & if\ Y_{xyc}=1 \\
(1-Y_{xyc})^\beta (\hat{Y}_{xyc})^\alpha \log\left(1-\hat{Y}_{xyc}\right) & otherwise
\end{cases}
\end{aligned}
$$

$$(1)$$

where $\alpha$ and $\beta$ are hyper-parameters of the focal loss, and $N$ is the number of center points in input $I$. The normalization by $N$ is chosen to normalize all positive focal loss instances to 1. We use $\alpha = 2$ and $\beta = 4$ in all our experiments, following Law and Deng [20].

*2) Box Regression:* The bounding box regression module only happens on the positive center point features. For each bounding box, an eight-dimensional vector $\left[ d_x, d_y, z, l, w, h, \cos(r), \sin(r) \right]$ is regressed to represent the object instance in LiDAR point clouds. $d_x$, $d_y$ are the discretization deviations of center points on the last feature map, $z$ is the absolute coordinate value in $Z$ axis, $l$, $w$, $and$, $h$ are the 3D sizes: length, width, and height, $\cos(r)$ and $sin(r)$ are trigonometric function values of rotation angle $r$ around $Z$ axis. Thus, box regression includes offset regression, z-coord regression, size regression, and direction regression.

*a) Offset regression:* To recover the 2D discretization error $\frac{p}{R} - \tilde{p}$ caused by the output stride, the offset regression is used to predict an offset feature map $\hat{O} \in \mathcal{R}^{\frac{L}{R} \times \frac{W}{R} \times 2}$ for each center point. All classes $c$ share the same offset prediction. Because the ground truth offsets fall between 0 and 1. A logistic activation is used to constrain the offset predictions to fall in this range. The offset is trained with an L1 loss:

$$L_{off} = \frac{1}{N} \sum_p \sum_{i \in (\Delta x, \Delta y)} \left| \sigma(\hat{O}_{\tilde{p},i}) - O_{p,i} \right| \quad (2)$$

*b) Direction regression:* To predict the rotation angle around $Z$ axis and solve the adversarial example problem between the cases of 0 and $\pi$ radians, we encode each rotation angle $r$ as $(cos(r), sin(r))$, and decode the $r$ as atan2 $(sin(r), cos(r))$ during inference. The rotation angle is factored into two correlated values. Thus, the direction regression predicts a direction feature map $\hat{D} \in \mathcal{R}^{\frac{L}{R} \times \frac{W}{R} \times 2}$ for each center point. The direction is also trained with an L1 loss:

$$L_{dir} = \frac{1}{N} \sum_p \sum_{i \in \{\sin(r), con(r)\}} \left| \sigma(\hat{D}_{\tilde{p},i}) - D_{p,i} \right| \quad (3)$$

*c) Z-coord regression:* The z-coord regression is used to predict the center location of the bounding box in $Z$ axis. It outputs a z-coord feature map $\hat{Z} \in \mathcal{R}^{\frac{L}{R} \times \frac{W}{R} \times 1}$ for each center point. All classes $c$ share the same z-coord prediction. However, owing to the unbounded regression targets, the model is sensitive to outliers. These outliers, which can be regarded as hard samples, will produce excessively large gradients that are harmful to the training process. The inliers, which can be regarded as the easy samples, contribute little gradient to the overall gradients compared with the outliers. Considering this issue, the balanced L1 loss proposed in [32] is employed for training z-coord regression:

$$L_z = \frac{1}{N} \sum_p L_b\left(\left|\hat{Z}_{\hat{p}} - Z_p\right|\right) \tag{4}$$

where, $L_b$ is balanced L1 loss:

$$L_b(x) = \begin{cases} \frac{a}{b}(b|x|+1)\ln(b|x|+1) - a|x| & if \ |x| < 1 \\ \gamma |x| + C & otherwise \end{cases}$$

in which $a$, $b$ and $\gamma$ are hyper-parameters of balanced L1 loss, and they are constrained by

$$a\ln(b+1) = \gamma$$

We use $a = 0.5$ and $\gamma = 1.5$ in all our experiments, following [32].

*d) Size regression:* To predict the length $l$, width $w$ and height $h$, the size regression is employed. To limit the computational burden, we regress a single size for all classes. The size regression outputs a size feature map $\hat{S} \in \mathcal{R}^{\frac{L}{R} \times \frac{W}{R} \times 3}$. Like z-coord regression, the size regression also introduces training imbalance owing to the unbounded regression targets. So, the balanced L1 loss proposed in [32] is also employed for training size regression:

$$L_{size} = \frac{1}{N} \sum_p \sum_{i \in \{l, w, h\}} L_b \left|\hat{S}_{\hat{p}, i} - S_{p, i}\right| \tag{5}$$

*3) Corner Classification:* Unlike 2D images there often exists a pixel near the object center, as LiDARs only capture the surfaces of objects, 3D object centers are likely to be in empty space. Besides, center points may be far from the boundary of objects, so they are difficult to estimate accurate boundaries. We want our model to learn the object shape and structure information, so we introduce another supervision signal for corner classification. Like center classification, the corner classification module is used to classify four corner points of bounding boxes in $XY$ plane. It outputs corner heatmaps and each corner heatmap corresponds to one category, let $\hat{A} \in [0, 1]^{\frac{L}{R} \times \frac{W}{R} \times C}$ be the corner heatmaps, where $R$ is the downsample stride and $C$ is the number of classes. A prediction $\hat{A}_{x,y,c} = 1$ corresponds to a detected corner point, while $\hat{A}_{x,y,c} = 0$ is background. We generate ground truth corner heatmaps $A \in [0, 1]^{\frac{L}{R} \times \frac{W}{R} \times C}$ following the above center classification. The training corner loss $L_{cor}$ is also a penalty-reduced focal loss similar to center classification.

*4) Decode Loss:* In the above box regression, the different properties of bounding boxes are considered to be independent of each other, and they are predicted by different modules in the detection head. To help the detector learn the implicit relationship between the different properties of 3D bounding boxes, the decode loss is proposed.

In the training stage, the outputs of bounding box regression are first decoded into eight corners, and then balanced L1 loss is computed on the coordinates of eight corners directly with regard to ground truth. Since the decoding from regression outputs to eight corners is just a combination of some normal mathematic operations, this decoding process is differentiable, and gradients can be back-propagated through this decoding process. Thus, we believe that this decoding loss provides explicit supervision for the relationship between different properties of 3D bounding boxes. The decode loss is the distance between the decoded eight corners and corresponding ground truth, expressed as

$$L_{decode} = \frac{1}{N} \sum_p \sum_c L_b\left(\left\|\hat{P}_c - P_c\right\|\right) \tag{6}$$

where $\hat{P}_c$ and $P_c$ are the coordinate of prediction and ground truth for corner $c$.

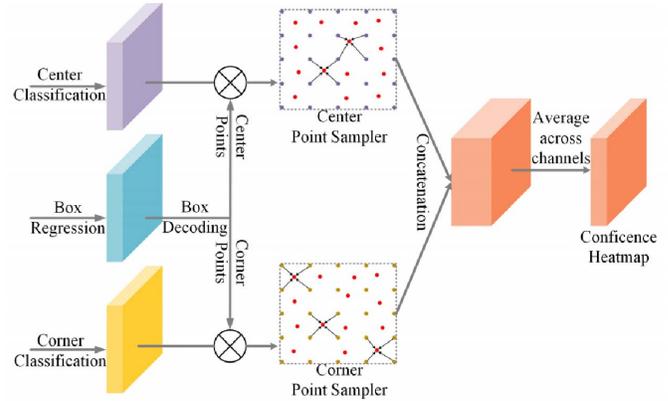

Fig. 2. Keypoint-sensitive warping. We generate 5 sampling points from each bounding box predicted by box regression branch. Each keypoint is used to sample the feature point from the corresponding heatmap using a bilinear interpolation kernel. The final confidence map is computed by taking the average among 5 sampled maps.

### B. Keypoint-Sensitive Warping

To solve the misalignment between the predicted bounding boxes and corresponding confidence maps, we propose a keypoint-sensitive warping operation, namely KSWarp, as an efficient variant of PSWarp [15] and PSRoIAlign [33], to align the classification confidences with the predicted bounding boxes by performing spatial transformation on the feature maps. Different from PSRoIAlign and PSWarp, our KSWarp does not participate at the training stage and it only operates at inference stage.

At the inference stage, we decode each predicted bounding box into 5 keypoints in the $XY$ plane: a center point and four corner points. Each keypoint encodes information of a certain part of the object, {center, upper-left, upper-right, bottom-left, bottom-right}. Then, we choose the decoded center points as sampling points in center heatmaps $\hat{Y}$, and the corner points as sampling points in the corner heatmaps $\hat{A}$. In this way, we can get 5 sampling keypoints and each of which associates with the corresponding heatmap. Our KSWarp consists of two feature samplers: center point sampler and corner point sampler. As shown in Fig. 2, the center point sampler takes the center heatmaps and the decoded center points as input, producing center maps sampled from the center heatmaps at center points. The corner point sampler takes the corner heatmaps and the decoded corner points as input, producing corner maps sampled from the corner heatmaps at corner points. The final confidence map $C$ is computed by taking the average among 5 sampled maps. Given a predicted bounding box $p$ and its corresponding sample keypoints $\{(u^k, v^k) = S_p^k : k = 1, 2, \cdots, 5\}$, the final confidence of this bounding box can be calculated by:

$$C_p = \frac{1}{5} \sum_{k=1}^{5} \sum_{\substack{i \in \left\{ \lfloor u^k \rfloor, \lfloor u^k + 1 \rfloor \right\} \\ j \in \left\{ \lfloor v^k \rfloor, \lfloor v^k + 1 \rfloor \right\}}} \chi_{ij}^k \times b\left(i, j, u^k, v^k\right) \tag{7}$$

where $\chi^k$ is the corresponding keypoint heatmap, $b$ is a bilinear sampling kernel which has a form of $b(i, j, u, v) = max(1 - |i - u|, 0) \times max(1 - |j - v|, 0)$.

Compared to PSRoIAlign [33], KSWarp is more efficient since it mitigates the need of generating RoIs from a dense feature map with NMS. It considers only 5 keypoints for each cell in box regression map. Therefore, it has the same computational complexity as a standard convolution. In PSWarp [15] module, the bounding box is divided into several sub-windows with ambiguous meanings, and then the center position is selected as the sampling point. Compared with PSWarp, the proposed KSWarp directly uses 5 keypoints with apparent appearance characteristics for sampling. On the one hand, these 5 keypoints provide explicit localization information. On the other hand, the center and corner features are more discriminative, and they can be easily learned through the corner classification and the center classification module. The above two aspects enable KSWarp to achieve better consistency in localization and classification. In addition, the number of sampling points of the proposed KSWarp is greatly reduced which makes our KSWarp more lightweight and efficient than PSWarp. And our KSWarp only runs during the inference stage, and it can be used as a post-processing module at negligible cost.

### C. Training and Inference

The final loss for our proposed CenterNet3D is the weighted sum of all above classification loss and regression losses:

$$L = \delta_{cls} L_{cls} + \delta_{off} L_{off} + \delta_z L_z + \delta_{size} L_{size} + \delta_{dir} L_{dir} + \delta_{cor} L_{cor} + \delta_{decode} L_{decode}$$

where, $\delta_{cls}, \delta_{off}, \delta_z, \delta_{size}, \delta_{dir}, \delta_{cor}$ and $\delta_{decode}$ are the weights to balance the center classification, offset regression, z-coord regression, size regression, direction regression corner classification loss, and decode loss.

At inference time, we first decode the predicted bounding boxes into center points and corner points, and they are used as the inputs of KSWarp module to obtain the final confidence heatmaps. Then, we extract the center predictions in confidence heatmaps for each class independently. We filter all confidence predictions by whether whose values are greater or equal to their 8-connected neighbors and only keep those confidence predictions whose values are above the predefined threshold as detected center points.

Let $\hat{P}_c = \{\hat{x}_i, \hat{y}_i\}_{i=1}^{n}$ be the set of $n$ detected center points of class $c$. $(\hat{x}_i, \hat{y}_i)$ is the integer coordinates in confidence heatmaps. $(\delta \hat{x}_i, \delta \hat{y}_i)$, $\hat{z}_i$, $(\hat{l}_i, \hat{w}_i, \hat{h}_i)$, $(\hat{sin}_i, \hat{cos}_i)$ are corresponding offset, z-coord, size and direction prediction at location $(\hat{x}_i, \hat{y}_i)$. We use the prediction of KSWarp as a measure of its detection confidence, and produce a bounding box: $(\hat{x}_i + \delta \hat{x}_i, \hat{y}_i + \delta \hat{y}_i, \hat{z}_i, \hat{l}_i, \hat{w}_i, \hat{h}_i, atan2(\hat{sin}_i, \hat{cos}_i))$ All outputs are produced directly from the confidence heatmaps without the need for non-maximum suppression or other post-processing. The confidence prediction filter serves as a sufficient non-maximum suppression alternative and can be implemented efficiently using a $3 \times 3$ max pooling operation.

## IV. EXPERIMENTS

In this section, we evaluate our method on two common 3D object detection datasets, including the KITTI dataset [49] and a larger and more complex nuScenes dataset [50]. In the following, we first present a brief introduction to these datasets in Section A. Then, we introduce the implementation details of our method in Section B. In Section C we exhibit the comparisons with state-of-the-art methods on the KITTI dataset and the nuScenes dataset. In Section E, we present ablation studies about our method.

### A. Dataset and Evaluation

The KITTI dataset is the widely used dataset for evaluating 3D object detectors. The dataset contains 7,481 annotated LiDAR frames for training with 3D bounding boxes for object classes such as *cars*, *pedestrians*, and *cyclists*. Following the common protocol, we further divide the training data into a training set with 3,712 frames and a validation set with 3,769 frames. The detection task is divided into three different levels of difficulty: *easy*, *moderate*, and *hard* based on the object size, occlusion state, and truncation level. We conduct experiments on the most commonly used *car* and *pedestrian* categories. All results are evaluated by the mean average precision with a rotated IoU threshold 0.7 for *car* and 0.5 for *pedestrian*. The average precision (AP) is calculated using 40 recall positions.

The nuScenes dataset is a more challenging dataset. It contains 1000 scenes, gathered from Boston and Singapore, due to their dense traffic and highly challenging driving situations. It provides us with 1.4M 3D objects on 10 different classes, as well as their attributes and velocities. To predict velocity and attribute, all former methods combine points from key frames and frames in last 0.5s. In the benchmark, a new evaluation metric called nuScenes detection score (NDS) is presented, which is a weighted sum between mean average precision (mAP), the mean average errors of location (mATE), size (mASE), orientation (mAOE), attribute (mAAE) and velocity (mAVE). We use TP to denote the set of the 5 mean average errors, and NDS is calculated by

$$NDS = \frac{1}{10} \left[ 5mAP + \sum_{mTP \in TP} (1 - min(1, mTP)) \right] \tag{8}$$

### B. Implementation Details

*1) Backbone Network:* For the KITTI benchmark, we crop the point cloud based on the ground-truth distribution in $[-3, 1] \times [-40, 40] \times [0, 70]m$ along the $Z$, $Y$, $X$ axes. We use a voxel size of $V_x = V_y = 0.05m$, $V_z = 0.1m$ to voxelize the cropped point cloud into a regular grid map. For the nuScenes dataset, the range of point cloud is within $[-5, 3] \times [-54, 54] \times [-54, 54]$m in $Z, Y, X$ axes which is voxelized with voxel size of $V_x = V_y = 0.075m$, $V_z = 0.2m$.

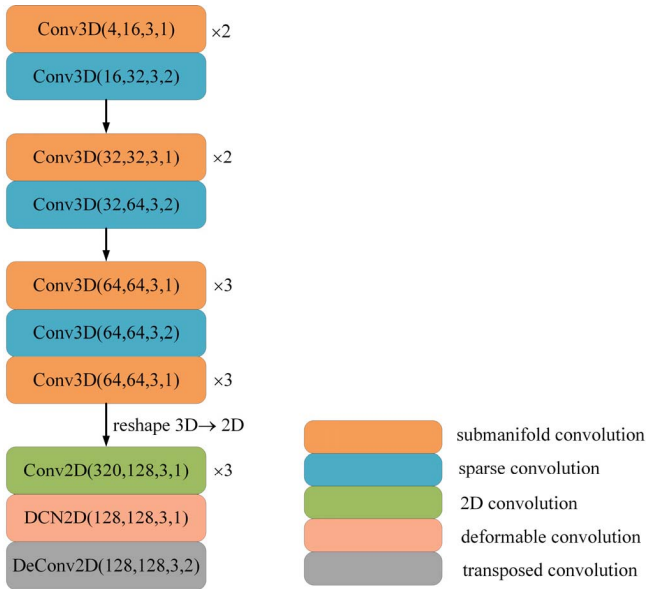

Fig. 3. The details of backbone network for CenterNet3D. Conv (cin, cout, k, s) represents a convolutional block, where cin, cout, k, s denotes input channel number, output channel number, kernel size and stride respectively. Each block consists of Convolution, BatchNorm, Leaky ReLU. The stride of each 3D sparse convolution is 2 which is used to downsample the feature maps.

In regular grid map, the car and pedestrian instance only occupies 30-60 and 12-16 voxels respectively, which are typically small objects. To improve car detection performance, we apply transposed convolution layer and deformable convolution layer to obtain high resolution and strong semantic feature maps.

The details of our architecture of the backbone network are shown in Fig. 3. We use the simplified version of Voxel Feature Encoder, i.e. VFE [8], by taking the mean of a fixed number of points sampled in a voxel. Our backbone network has 3D and 2D parts. The 3D part has 10 submanifold convolution blocks and 3 sparse convolution blocks. The 2D part has 3 convolution blocks, 1 deformable convolution block, and 1 transposed convolution block. We downsample in total three times with sparse convolution blocks, then the 3D feature maps are transformed to 2D by collapsing the height dimension. Finally, the 2D feature maps are processed by deformable convolution block and transposed convolution block to obtain high-resolution and strong semantic feature maps for CenterNet3D head.

Since the output feature maps of the backbone network are collapsed to bird eye view, we thus in this paper assign center points and corner points in bird eye view. For CenterNet3DSplit, the detection head has six modules (center classification, offset regression, z-coord regression, size regression, direction regression and corner classification), for CenterNet3DMerge, the detection head has three modules (center classification, box regression and corner classification). All sub-task modules share the common backbone network.

For each module, the features of the backbone are passed through a separate $3 \times 3$ convolution, Leaky ReLU, and another $1 \times 1$ convolution. For center classification, corner classification, and offset regression, another logistic activation is used to constrain the predictions in between 0 and 1. For direction regression, another hyperbolic tangent activation is used to constrain the prediction in between $-1$ and 1. Compared with anchor-based methods, our method only has one positive sample for each ground truth which leads to an extreme imbalance of positive and negative samples. To mitigate the imbalance, we set $t = 0.01$ to obtain more non-zero pixels in ground truth center heatmap $Y \in [0, 1]^{\frac{L}{R} \times \frac{W}{R} \times C}$ and corner heatmap $A \in [0, 1]^{\frac{L}{R} \times \frac{W}{R} \times C}$. We set $\delta_{cls} = \delta_{decode} = 0.5$, $\delta_{cor} = 0.1$, $\delta_{off} = \delta_z = \delta_{size} = \delta_{dir} = 1$ in the weighted sum of losses.

*2) Training and Inference:* The network is trained by ADAM [34] optimizer with fixed weight decay 0.01. The learning schedule is one-cycle policy [35] with maximum learning rate 2.25e-3, division factor 10, momentum ranges from 0.95 to 0.85. The network is trained with batch size 6 for 80 epochs. During the inference, a $3 \times 3$ max pooling operation is applied on the confidence heatmap, then we keep 50 predictions whose confidences are larger than 0.1. For the nuScenes dataset, we combine key frame with frames within last 0.5s so as to predict velocity and attribute, just the same as other methods.

*3) Data Augmentation:* For the KITTI dataset, we adopt 4 different data augmentation strategies. First, we perform a common cut-and-paste strategy [12], [15] for data augmentation. Specifically, we randomly add foreground instances with their inner points from other frames to current point cloud. Each instance is followed by a collision test to avoid the violation of the physical rule. All ground-truth boxes are individually augmented. Then, each box is randomly rotated and translated. The noise for the rotation is uniformly drawn from $[-\frac{\pi}{4}, \frac{\pi}{4}]$ and the noise for the translation is drawn from $\mathcal{N}(0, 1)$, $\mathcal{N}(0, 1)$ and $\mathcal{N}(0, 1)$ for $X, Y, Z$ respectively. Finally, we apply random flipping, global rotation, and global scaling to the whole point cloud. The noise for global rotation is uniformly drawn from $[-\frac{\pi}{4}, \frac{\pi}{4}]$ and the scaling factor is uniformly drawn from $[0.95, 1.05]$.

For the data augmentation of nuScenes dataset, we use the same augmentation strategies including the random flip, global rotation, and global scaling with the same parameters. Note that cut-and-paste and per-object augmentation strategies are not applied because of the sufficient scale of nuScenes dataset to prevent the overfitting.

## C. Experimental Results on KITTI Test Set

We compare our CenterNet3D detector with other state-of-the-art approaches by submitting the *car* and *pedestrian* detection results to the KITTI server for evaluation. As shown in Table I, we evaluate our method on the 3D benchmark and the birds eye view benchmark. As we can see, our proposed CenterNet3DSplit surpasses most of the methods in Table I, except for PV-RCNN [31], RangeRCNN [25], STD [26] and SASSD [15]. And our CenterNet3DSplit

TABLE I

PERFORMANCE COMPARISON WITH PREVIOUS METHODS ON KITTI TEST SERVER FOR CAR AND PEDESTRIAN DETECTION. THE BOLD VALUE INDICATES THE TOP PERFORMANCE FOR ONE STAGE AND TWO STAGE METHODS

| Method | Modality | Stage | 3D Detection (Car) | | | Bev Detection (Car) | | | 3D Detection (Peds.) | | | Bev Detection (Ped.) | | | FPS |
|---|---|---|---|---|---|---|---|---|---|---|---|---|---|---|---|
| | | | Easy | Mod. | Hard | Easy | Mod. | Hard | Easy | Mod. | Hard | Easy | Mod. | Hard | |
| AVOD-FPN [52] | L+C | Two | 83.07 | 71.76 | 65.73 | 90.99 | 84.82 | 79.62 | 50.46 | 42.27 | 39.04 | 58.75 | **51.05** | 47.54 | 10 |
| PointRCNN [24] | L | Two | 86.96 | 75.64 | 70.70 | 92.13 | 87.39 | 82.72 | 47.98 | 39.37 | 36.01 | 54.77 | 46.13 | 42.84 | 10 |
| F-ConvNet [42] | L+C | Two | 87.36 | 76.39 | 66.69 | 91.51 | 85.84 | 76.11 | 52.16 | **43.38** | 38.80 | 57.04 | 48.96 | 44.33 | 2.1 |
| MMF [53] | L+C | Two | 88.40 | 77.43 | 70.22 | 93.67 | 88.21 | 81.99 | - | - | - | - | - | - | **12.5** |
| STD [26] | L | Two | 87.95 | 79.71 | 75.09 | 94.74 | 89.19 | **86.42** | 53.29 | 42.47 | 38.35 | **60.02** | 48.72 | 44.55 | 10 |
| RangeRCNN [25] | L | Two | 88.47 | 81.33 | 77.09 | 92.15 | 88.40 | 85.74 | - | - | - | - | - | - | 12.5 |
| PV-RCNN [31] | L | Two | **90.25** | **81.43** | **76.82** | 94.98 | 90.65 | 86.14 | 52.17 | 43.29 | **40.29** | 59.86 | 50.57 | 46.74 | 8 |
| SECONDv1.5 [12] | L | One | 84.65 | 75.96 | 68.71 | 91.81 | 86.37 | 81.04 | 51.07 | 42.56 | 37.29 | 55.10 | 46.27 | 44.74 | 20 |
| TANet [54] | L | One | 84.39 | 75.94 | 68.82 | 91.58 | 86.54 | 81.19 | **53.72** | 44.34 | 40.49 | **60.85** | **51.38** | 47.54 | 25 |
| PointPillars [13] | L | One | 82.58 | 74.31 | 68.99 | 90.07 | 86.56 | 82.81 | 51.45 | 41.92 | 38.89 | 57.60 | 48.64 | 45.78 | **42** |
| HotSpotNet [36] | L | One | 88.12 | 78.34 | 73.49 | 94.06 | 88.09 | 83.24 | 53.10 | **45.37** | 41.47 | 57.39 | 50.53 | 46.65 | 20 |
| SASSD [15] | L | One | **88.75** | **79.79** | 74.16 | **95.03** | **91.03** | 85.96 | - | - | - | - | - | - | 25 |
| CenterNet3DMerge | L | One | 87.68 | 78.55 | 72.57 | 90.26 | 88.29 | 85.71 | 52.56 | 43.66 | 41.21 | 57.23 | 49.59 | 46.55 | 26 |
| CenterNet3DSplit | L | One | 88.23 | 79.23 | **75.34** | 92.74 | 88.51 | **86.32** | 53.32 | 44.42 | **41.63** | 57.46 | 50.25 | **47.83** | 25 |

and CenterNet3DMerge can both achieve competitive performance with other anchor-based methods, the CenterNet3DSplit achieves performance comparable to STD [26] and SASSD [15] and it shows better performance on hard level where objects are usually far away, occluded, and truncated.

Note that all other methods listed in Table I are anchor-based except HotSpotNet [36]. HotSpotNet is an anchor-point based 3D detectors. Although it does not require predefined anchors, it assigns multiple anchor points to one object. So it also requires heavy NMS post-processing which is hard to differentiate and train. PV-RCNN, RangeRCNN, STD and SASSD are also anchor-based methods, they both need to place a large number of anchor boxes to overlap different aspect ratios and scales which brings extra hyper-parameters and NMS post-processing. Moreover, RangeRCNN, STD and PV-RCNN are both two-stage methods, STD and PV-RCNN both used PointNet-based backbone which is stronger than the voxel-based backbone used in our CenterNet3D. Besides, PV-RCNN has a stronger pipeline, it uses complex Voxel Set Abstraction module to integrate the advantages of voxel-based and point-based networks.

In particular, our method does not need predefined anchors which simplifies the design choices of the model and reduces unnecessary hyperparameters. The proposed CenterNet3D assigns the "anchor" based solely on location, So our method does not need NMS post-processing. The proposed CenterNet3DMerge performs slightly worse than CenterNet3DSplit. It is demonstrated that different modules for different sub-tasks can be more specialized, they can learn the specific appearance characteristics of an object instance. In the rest of the paper, without further emphasizing, we will adopt CenterNet3DSplit as our method for quantitative evaluations.

In addition, we also compared with the model AFDet [23], which is also a keypoint based 3D detector. Both our method and AFDet use anchor-free detection head for center classification and 3D bounding box regression. However, to alleviate the impact of the center of bounding box in empty space, we propose an extra corner attention module to help the network be aware of object size and structure information. Moreover, we design a lightweight post-processing module

TABLE II

THE KITTI VALIDATION SET CAR DETECTION PERFORMANCE COMPARISON BETWEEN AFDET AND CENTERNET3D

| Method | 3D Detection (Car) | | | Bev Detection (Car) | | |
|---|---|---|---|---|---|---|
| | Easy | Mod. | Hard | Easy | Mod. | Hard |
| AFDet [23] | 85.68 | 75.57 | 69.31 | 89.42 | 85.45 | 80.56 |
| CenterNet3DSplit | **86.27** | **76.45** | **71.11** | **89.53** | **86.23** | **82.50** |

KSWarp to alleviate the misalignment between the localization accuracy and classification confidence. Since AFDet did not publish its performance in test set, we only compared our CenterNet3D with it in validation set. For a fair comparison, we use the point cloud encoder and 2D backbone network with the same configuration in [23]. The result of AFDet comes from the official release in [23]. As shown in Table II, our CenterNet3DSplit with the same backbone with AFDet achieves better performance on all metrics.

We can conclude that our proposed one-stage anchor free and NMS free CenterNet3D achieves competitive or even better results. The inspiring results show the success of representing 3D objects as center points as well as potentials of keypoint based anchor-free detectors in 3D detection.

We also visualize some prediction results for *car* and *pedestrian* on the validation set in Fig. 4, and we project the 3D bounding boxes detected from LiDAR to the RGB images for better visualization. The predicted bounding boxes are shown in green. The ground truth bounding boxes are shown in red. The word "FN" inside each box represents a false negative, the word "TP" represents a true positive, but the corresponding ground truth annotation is not provided because of heavy occlusion or invisibility in the camera view. However, these occluded or invisible objects are easily detected with 3D LiDAR. The word "FP" represents a false positive that is far away from the sensors and appears with sparse point clouds. In future work, we will focus on investigating ways to incorporate appearance cues from RGB images to prevent false positives.

### D. Experimental Results on NuScenes Test Set

We further conduct experiments on the nuScenes test dataset to verify the effectiveness of our approach in more complex multi-class object detection scenarios. For this purpose, we compare the proposed CenterNet3D with the other four methods: PointPillars,[1] SECOND[1], TANet,[2] SASSD,[3] which are anchor-based detectors. *Construction vehicle* and *bicycle* classes are omitted as the accuracies are too low to compare. The performances of competing methods are reproduced using their official code. We compare their APs of each class in Table III and show NDS and mAP among different methods in Table IV. As illustrated in Table IV, our method draws better performance compared to all anchor-based one-stage methods by a large margin. It outperforms PointPillars, SECOND, TANet and SASSD by 11.72%, 6.60%, 3.41% and 0.33% in terms of NDS. In particular, our method achieves the best performance on small targets, such as *pedestrian* and *traffic cone* and *motorcycle*, as shown in Table III. The results show that our model can deal with different objects with a large variance on scale without anchors.

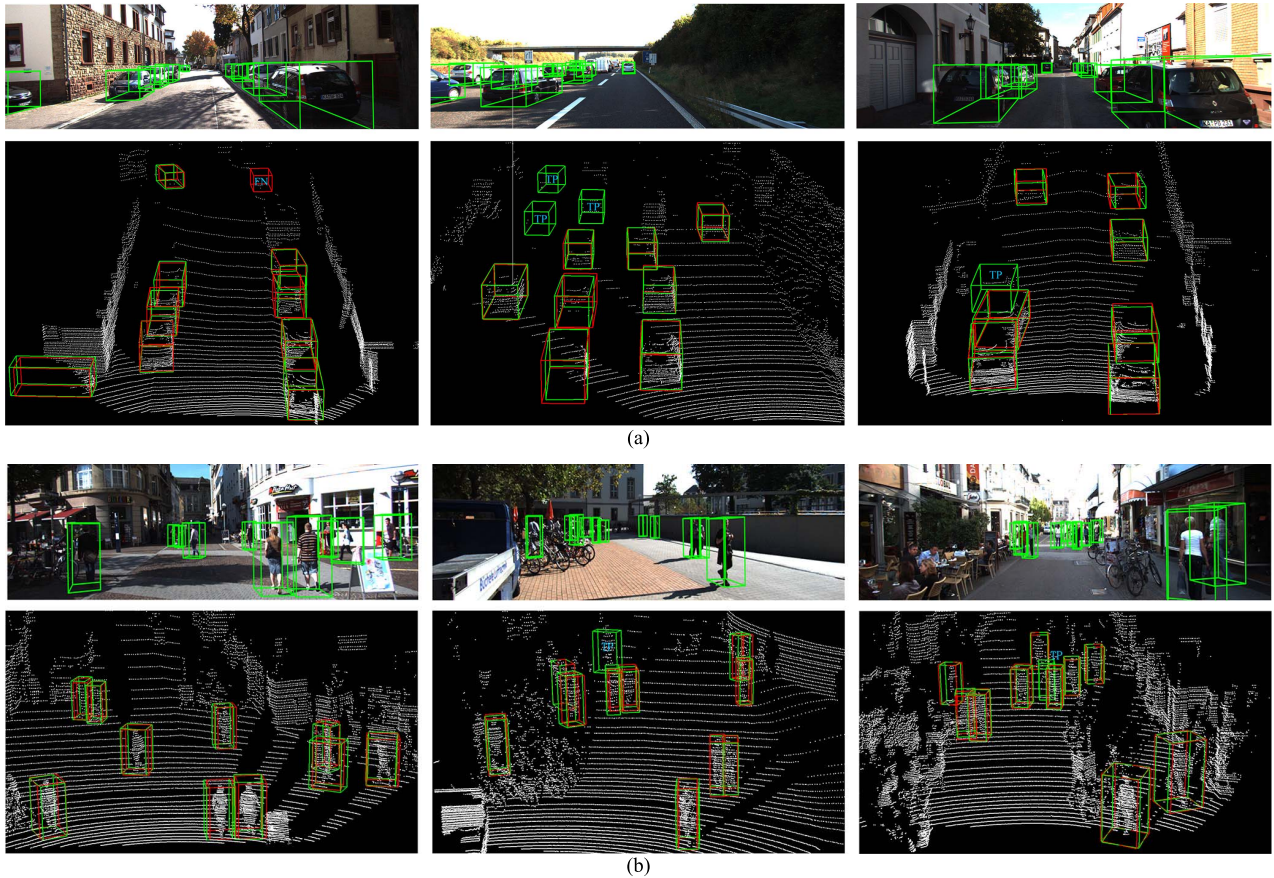

Fig. 4. Qualitative results of CenterNet3D for car and pedestrian detection on KITTI validation set. The predicted bounding boxes are shown in green. The ground truth bounding boxes are shown in red. The predictions are projected onto the RGB images (upper row) for better visualization. The word "FN" inside each box represents a false negative, the word "TP" represents a true positive, but the corresponding ground truth annotation is not provided because of heavy occlusion or invisibility in the camera view. The word "FP" represents a false positive.

Besides, we also visualize several prediction results for multi-classes on the nuScenes validation set in Fig. 5. Since the detection range is 360 degrees around the LiDAR, we only show the detection results under the LiDAR perspective.

*E. Ablation Studies*

In this section, we conduct an extensive ablation study to validate the effect of the ideas in the proposed CenterNet3D method. All models are trained on the training set and evaluated on the validation set of KITTI dataset for *Car* detection. All evaluations on the validation split are performed via 40 recall positions.

*1) Effects of Corner Attention Module:* To prove the effectiveness of the corner attention module, we remove the corner attention module and only use the center point sampler to obtain confidence heatmaps in KSWarp module. We show the results of our CenterNet3D with and without the corner attention module in Table IV. We can see that when our algorithm trained with the proposed corner attention module, the overall performance is boosted. Our proposed corner attention module yields an increase in the AP of 0.62%/0.67%/0.91% and 0.59%/0.62%/0.95% on easy/moderate/hard levels for 3D and BEV, respectively. It can be observed that the performance gains on hard level are higher than those on the easy and moderate levels. This is because the objects from hard level usually contain only a few points, the corner points are more distinguishing than center points. It is demonstrated that the proposed corner attention module is effective, and the corner features are beneficial to improve the discrimination of objects, especially for hard samples.

---

TABLE III
AP for Different Classes on nuScenes Test Set

| Methods | Car | Pedestrian | Bus | Barrier | Traffic Cone | Truck | Trailer | Motorcycle | mAP |
|---|---|---|---|---|---|---|---|---|---|
| PointPillars [13] | 68.40 | 59.72 | 28.58 | 38.91 | 30.80 | 23.00 | 23.40 | 27.40 | 37.53 |
| SECOND [12] | 76.25 | 64.47 | 32.12 | 40.45 | 45.60 | 27.74 | 26.56 | 30.60 | 42.97 |
| TANet [54] | 79.40 | 70.20 | 42.53 | **48.67** | 47.82 | 34.71 | 28.13 | 34.26 | 48.34 |
| SASSD [15] | **81.01** | 65.54 | **56.60** | 45.56 | 47.90 | **40.23** | 35.17 | 33.56 | 50.70 |
| CenterNet3DSplit | 80.23 | **70.58** | 55.26 | 47.32 | **48.21** | 38.79 | **37.83** | **34.40** | **51.57** |

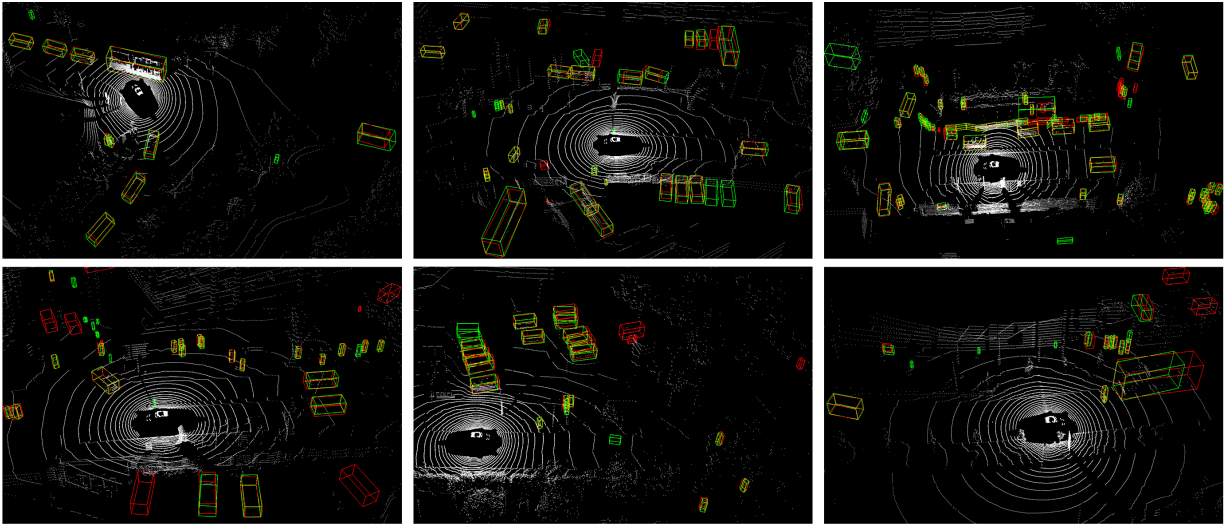

Fig. 5. Qualitative results of CenterNet3D on nuScenes validation set. The predicted bounding boxes are shown in green. The ground truth bounding boxes are shown in red.

TABLE IV
NDS on nuScenes Test Dataset

| Methods | mAP | mATE | mASE | mAOE | mAVE | AAE | NDS |
|---|---|---|---|---|---|---|---|
| PointPillars [13] | 37.53 | 0.43 | 0.26 | 0.39 | 0.35 | 0.40 | 50.47 |
| SECOND [12] | 42.97 | 0.40 | 0.23 | 0.32 | 0.29 | 0.35 | 55.58 |
| TANet [54] | 48.34 | 0.38 | 0.19 | 0.34 | 0.27 | 0.36 | 58.77 |
| SASSD [15] | 50.70 | 0.30 | **0.18** | **0.29** | **0.24** | 0.34 | 61.85 |
| CenterNet3DSplit | **51.57** | **0.29** | 0.20 | 0.30 | 0.25 | **0.32** | **62.18** |

TABLE V
Comparison With PSWarp AND PSRoIAlign on Kitti Validation Set for "Car" Detection

| Method | 3D Detection (Car) | | | Time |
|---|---|---|---|---|
| | Easy | Mod. | Hard | |
| PSRoIAlign [33] | **90.13** | **80.02** | 78.07 | ~5ms |
| PSWarp [15] | 89.79 | 79.68 | 78.01 | ~1ms |
| KSWarp | 89.91 | 79.48 | **78.11** | ~0.3ms |

*2) Effects of Keypoint-Sensitive Warping:* To prove the effectiveness of the proposed keypoint-sensitive warping module, we remove the KSWarp module and extract the center predictions in center heatmaps for each class independently. We show the performance of CenterNet3D with and without KSWarp in Table VII. We can see that the proposed KSWarp can improve the performance by 0.53%/0.58%/0.67% and 0.51%/0.42%/0.79% on easy/moderate/hard levels for 3D and BEV, respectively. This validates the effectiveness of refining the classification confidences to the predicted bounding boxes. Therefore, by considering the center confidence and corner confidence at the same time, the proposed KSWarp module can make the classification confidence more consistent with the localization.

We also compare KSWarp with its counterparts by replacing KSWarp with PSWarp and PSRoIAlign and retraining the model. For PSWarp and PSRoIAlign, the spatial resolution for bounding box alignment is 4 × 7. As can be seen from Table VII, our KSWarp exhibits comparable performance with PSRoIAlign and PSWarp, and KSWarp takes less runtime. The main reason is that PSRoIAlign needs to enumerate RoIs from a dense feature map, PSWarp needs to sample 28 points for each bounding box, and our KSWarp only needs to sample

5 more discriminative keypoints. Moreover, our KSWarp does not participate at the training stage and it only operates at inference stage.

*3) Effects of Decode Loss:* To further prove the effectiveness of the proposed decode loss, we remove the decode loss in training stage and show the results of our CenterNet3D in Table VIII. We can see that the proposed decode loss can improve the performance by 0.39%/0.28%/0.43% and 0.52%/0.47%/0.36% on easy/moderate/hard levels for 3D and BEV, respectively. Thus, the proposed decode loss can help the detector learn the implicit relationship between the different properties of 3D bounding boxes and further improve the performance.

*4) Effects of Box Size Loss:* In this part, we evaluate the performance of our approach when using different types of box size loss, including L1 loss, smooth L1 [48], and balanced L1 loss [32]. As can be seen from Table IX, compared with smooth L1 loss, the balanced L1 loss can improve the performance by 1.77%/1.64%/1.54% and 1.89%/2.05%/2.13% on easy/moderate/hard levels for 3D and BEV, respectively. And the balanced L1 loss achieved considerably the best performance. It is demonstrated that the balance L1 loss is effective for balancing the gradients of hard and easy samples.

 

TABLE VI

Effect of Corner Classification Module on KITTI Validation Set for "Car" Detection

| Method | 3D Detection | | | Bev Detection | | |
|---|---|---|---|---|---|---|
| | Easy | Moderate | Hard | Easy | Moderate | Hard |
| w/o corner classification | 89.23 | 78.74 | 77.56 | 92.83 | 89.68 | 88.59 |
| w corner classification | **89.91/+0.68** | **79.48/+0.74** | **78.11/+0.55** | **93.45/+0.62** | **90.31/+0.63** | **89.28/+0.69** |

TABLE VII

Effects of Keypoint-Sensitive Warping Module on KITTI Validation Set for "Car" Detection

| Method | 3D Detection | | | Bev Detection | | |
|---|---|---|---|---|---|---|
| | Easy | Moderate | Hard | Easy | Moderate | Hard |
| w/o KSWarp | 89.38 | 78.90 | 77.44 | 92.94 | 89.89 | 88.49 |
| w KSWarp | **89.91/+0.53** | **79.48/+0.58** | **78.11/+0.67** | **93.45/+0.51** | **90.31/+0.42** | **89.28/+0.79** |

TABLE VIII

Effect of Decode Loss on KITTI Validation Set for "Car" Detection

| Method | 3D Detection | | | Bev Detection | | |
|---|---|---|---|---|---|---|
| | Easy | Moderate | Hard | Easy | Moderate | Hard |
| w/o decode loss | 89.52 | 79.20 | 77.68 | 92.93 | 89.84 | 88.92 |
| w decode loss | **89.91/+0.39** | **79.48/+0.28** | **78.11/+0.43** | **93.45/+0.52** | **90.31/+0.47** | **89.28/+0.36** |

TABLE IX

A Comparison of the Performances of Different Box Size Regression Loss on the KITTI Validation Set for "Car" Detection

| Method | 3D Detection | | | Bev Detection | | |
|---|---|---|---|---|---|---|
| | Easy | Moderate | Hard | Easy | Moderate | Hard |
| Smooth L1 Loss | 88.14 | 77.84 | 76.57 | 91.56 | 88.26 | 87.15 |
| L1 Loss | 89.56+1.42 | 79.12+1.28 | 77.89+1.32 | 93.01+1.45 | 89.74+1.48 | 88.56+1.41 |
| Balanced L1 Loss | **89.91+1.77** | **79.48+1.64** | **78.11+1.54** | **93.45+1.89** | **90.31+2.05** | **89.28+2.13** |

TABLE X

A Comparison of the Performances of Angle Regression Loss on the KITTI Validation Set for "Car" Detection

| Method | 3D Detection | | | Bev Detection | | |
|---|---|---|---|---|---|---|
| | Easy | Moderate | Hard | Easy | Moderate | Hard |
| Residual-based Loss | 87.53 | 77.67 | 76.59 | 89.96 | 88.74 | 86.74 |
| Sine-error Loss | 88.56+1.03 | 78.34+0.67 | 77.76+1.17 | 91.97+2.01 | 89.76+1.02 | 87.52+0.78 |
| Bin-based Loss | 89.85+2.32 | **79.58+1.91** | **78.35+1.76** | **93.56+3.60** | 90.23+1.49 | 88.79+2.05 |
| Sin-cos loss | **89.91+2.38** | 79.48+1.81 | 78.11+1.52 | 93.45+3.49 | **90.31+1.57** | **89.28+2.54** |

*5) Effects of Direction Loss:* We also compare the performances when using different types of direction loss in training stage, which include sine-error Loss [12], bin-based loss [24], residual-based loss [8] and sin-cos loss [28] used in this paper. Here the residual-based loss only encodes residual of rotation angle by $\Delta\gamma$ which cannot eliminate the ambiguity of angle regression. Our experiments in Table X show that sin-cos loss and bin-based loss achieve approximatively excellent performance. However, bin-based loss contains complex encoding and decoding for bin classification and residual regression which is computationally inefficient.

## V. Conclusion

We propose a novel representation, object as the center point, and a one-stage anchor-free and NMS-free 3D object detector, CenterNet3D, for autonomous driving. Our CenterNet3D is a keypoint based 3D object detector that finds the center points of objects and directly regresses bounding boxes. To solve the sparsity of point clouds and the hollowness near the centers of objects, we propose a corner classification module to learn discriminative corner features, which makes the network aware of object size and structure information. To alleviate the misalignment between the localization accuracy and classification confidence, we design an efficient keypoint-sensitive warping operation for postprocessing. Extensive experiments have validated the effectiveness of the ideas in the proposed CenterNet3D. The proposed CenterNet3D is simple, fast, accurate, and end-to-end differentiable without NMS post processing. On the KITTI and nuScenes benchmark, our proposed CenterNet3D achieves competitive performance with other one-stage anchor-based methods which shows the potentials of keypoint based anchor-free detectors in 3D detection.

## References

[1] X. Chen, K. Kundu, Z. Zhang, H. Ma, S. Fidler, and R. Urtasun, "Monocular 3D object detection for autonomous driving," in *Proc. IEEE Conf. Comput. Vis. Pattern Recognit. (CVPR)*, Jun. 2016, pp. 2147–2156.

[2] A. Mousavian, D. Anguelov, J. Flynn, and J. Kosecka, "3D bounding box estimation using deep learning and geometry," in *Proc. IEEE Conf. Comput. Vis. Pattern Recognit.*, Apr. 2017, pp. 7074–7082.

[3] A. Simonelli, S. R. Bulo, L. Porzi, M. Lopez-Antequera, and P. Kontschieder, "Disentangling monocular 3D object detection," in *Proc. IEEE/CVF Int. Conf. Comput. Vis. (ICCV)*, Oct. 2019, pp. 1991–1999.

[4] B. Xu and Z. Chen, "Multi-level fusion based 3D object detection from monocular images," in *Proc. IEEE/CVF Conf. Comput. Vis. Pattern Recognit.*, Jun. 2018, pp. 2345–2353.

[5] X. Z. Chen, K. Kundu, Y. Zhu, S. Fidle, R. Urtasun, and H. Ma, "3D object proposals using stereo imagery for accurate object class detection," *IEEE Trans. Pattern Anal. Mach. Intell.*, vol. 40, no. 5, pp. 1259–1272, May 2018, doi: 10.1109/TPAMI.2017.2706685.

[6] P. Li, X. Chen, and S. Shen, "Stereo R-CNN based 3D object detection for autonomous driving," in *Proc. IEEE/CVF Conf. Comput. Vis. Pattern Recognit. (CVPR)*, Jun. 2019, pp. 7644–7652.

[7] Z. Qin, J. Wang, and Y. Lu, "Triangulation learning network: From monocular to stereo 3D object detection," in *Proc. IEEE Int. Conf. Comput. Vis.*, Jun. 2019, pp. 7607–7615.

[8] A. Palffy, J. Dong, J. F. P. Kooij, and D. M. Gavrila, "CNN based road user interaction using the 3D radar cube," *IEEE Robot. Autom. Lett.*, vol. 5, no. 2, pp. 1263–1270, Apr. 2020, doi: 10.1109/LRA.2020.2967272.

[9] O. Schumann, C. Wöhler, M. Hahn, and J. Dickmann, "Comparison of random forest and long short-term memory network performances in classification tasks using radar," in *Proc. Sensor Data Fusion, Trends, Solutions, Appl. (SDF)*, Oct. 2017, pp. 1–6, doi: 10.1109/SDF.2017.8126350.

[10] N. Scheiner, N. Appenrodt, J. Dickmann, and B. Sick, "Radar-based feature design and multiclass classification for road user recognition," in *Proc. IEEE Intell. Vehicles Symp. (IV)*, Jun. 2018, pp. 779–786, doi: 10.1109/IVS.2018.8500607.

[11] Y. Zhou and O. Tuzel, "VoxelNet: End-to-end learning for point cloud based 3D object detection," in *Proc. IEEE/CVF Conf. Comput. Vis. Pattern Recognit.*, Jun. 2018, pp. 4490–4499.

[12] Y. Yan, Y. Mao, and B. Li, "SECOND: Sparsely embedded convolutional detection," *Sensors*, vol. 18, no. 10, p. 3337, Oct. 2018.

[13] A. H. Lang, S. Vora, H. Caesar, L. Zhou, J. Yang, and O. Beijbom, "PointPillars: Fast encoders for object detection from point clouds," in *Proc. IEEE Conf. Comput. Vis. Pattern Recognit.*, May 2019, pp. 12697–12705.

[14] M. Ye, S. Xu, and T. Cao, "HVNet: Hybrid voxel network for LiDAR based 3D object detection," in *Proc. IEEE/CVF Conf. Comput. Vis. Pattern Recognit. (CVPR)*, Jun. 2020, pp. 1631–1640.

[15] C. He, H. Zeng, J. Huang, X.-S. Hua, and L. Zhang, "Structure aware single-stage 3D object detection from point cloud," in *Proc. IEEE/CVF Conf. Comput. Vis. Pattern Recognit.*, Jun. 2020, pp. 11870–11879.

[16] T.-Y. Lin, P. Dollar, R. Girshick, K. He, B. Hariharan, and S. Belongie, "Feature pyramid networks for object detection," in *Proc. IEEE Conf. Comput. Vis. Pattern Recognit. (CVPR)*, Jul. 2017, pp. 2117–2125.

[17] T.-Y. Lin, P. Goyal, R. Girshick, K. He, and P. Dollár, "Focal loss for dense object detection," in *Proc. IEEE Int. Conf. Comput. Vis. (ICCV)*, Oct. 2017, pp. 2980–2988.

[18] Z. Tian, C. Shen, H. Chen, and T. He, "FCOS: Fully convolutional one-stage object detection," in *Proc. IEEE/CVF Int. Conf. Comput. Vis. (ICCV)*, Oct. 2019, pp. 9627–9636.

[19] T. Kong, F. Sun, H. Liu, Y. Jiang, L. Li, and J. Shi, "FoveaBox: Beyond anchor-based object detector," 2019, *arXiv:1904.03797*. [Online]. Available: http://arxiv.org/abs/1904.03797

[20] H. Law and J. Deng, "Cornernet: Detecting objects as paired keypoints," in *Proc. Eur. Conf. Comput. Vis.*, 2018, pp. 734–750.

[21] X. Zhou, J. Zhuo, and P. Krähenbuhl, "Bottom-up object detection by grouping extreme and center points," in *Proc. IEEE Conf. Comput. Vis. Pattern Recognit.*, Jan. 2019, pp. 850–859.

[22] X. Zhou, D. Wang, and P. Krähenbühl, "Objects as points," 2019, *arXiv:1904.07850*. [Online]. Available: http://arxiv.org/abs/1904.07850

[23] R. Ge *et al.*, "AFDet: Anchor free one stage 3D object detection," 2020, *arXiv:2006.12671*. [Online]. Available: http://arxiv.org/abs/2006.12671

[24] S. Shi, X. Wang, and H. Li, "PointRCNN: 3D object proposal generation and detection from point cloud," in *Proc. IEEE/CVF Conf. Comput. Vis. Pattern Recognit. (CVPR)*, Jun. 2019, pp. 770–779.

[25] Z. Liang, M. Zhang, Z. Zhang, X. Zhao, and S. Pu, "RangeR-CNN: Towards fast and accurate 3D object detection with range image representation," 2020, *arXiv:2009.00206*. [Online]. Available: http://arxiv.org/abs/2009.00206

[26] Z. Yang, Y. Sun, S. Liu, X. Shen, and J. Jia, "STD: Sparse-to-dense 3D object detector for point cloud," in *Proc. IEEE Int. Conf. Comput. Vis.*, Jul. 2019, pp. 1951–1960.

[27] Z. Yang, Y. Sun, S. Liu, and J. Jia, "3DSSD: Point-based 3D single stage object detector," in *Proc. IEEE/CVF Conf. Comput. Vis. Pattern Recognit.*, Feb. 2020, pp. 11040–11048.

[28] B. Yang, W. Luo, and R. Urtasun, "PIXOR: Real-time 3D object detection from point clouds," in *Proc. IEEE Conf. Comput. Vis. Pattern Recognit.*, Jan. 2018, pp. 7652–7660.

[29] M. Simon, S. Milzy, K. Amende, and H.-M. Gross, "Complex-YOLO: An euler-regionproposal for real-time 3D object detection on point clouds," in *Proc. Eur. Conf. Comput. Vis.*, Mar. 2018, pp. 1–14.

[30] Y. Chen, S. Liu, X. Shen, and J. Jia, "Fast point R-CNN," in *Proc. IEEE Int. Conf. Comput. Vis.*, Aug. 2019, pp. 9775–9784.

[31] S. Shi *et al.*, "PV-RCNN: Point-voxel feature set abstraction for 3D object detection," in *Proc. IEEE/CVF Conf. Comput. Vis. Pattern Recognit.*, Jan. 2020, pp. 10529–10538.

[32] J Pang, K. Chen, J. Shi, H. Feng, W. Ouyang, and D. Lin, "Libra R-CNN: Towards balanced learning for object detection," in *Proc. IEEE Conf. Comput. Vis. Pattern Recognit.*, Jan. 2019, pp. 821–830.

[33] J. Dai, Y. Li, K. He, and J. Sun, "R-FCN: Object detection via region-based fully convolutional networks," 2016, *arXiv:1605.06409*. [Online]. Available: http://arxiv.org/abs/1605.06409

[34] I. Loshchilov and F. Hutter, "Decoupled weight decay regularization," 2017, *arXiv:1711.05101*. [Online]. Available: http://arxiv.org/abs/1711.05101

[35] L. N. Smith and N. Topin, "Super-convergence: Very fast training of neural networks using large learning rates," *Artif. Intell. Mach. Learn. Multi-Domain Oper. Appl.*, vol. 11006, May 2019, Art. no. 1100612.

[36] Q. Chen, L. Sun, Z. Wang, K. Jia, and A. Yuille, "Object as hotspots: An anchor-free 3D object detection approach via firing of hotspots," 2019, *arXiv:1912.12791*. [Online]. Available: http://arxiv.org/abs/1912.12791

[37] H. Kuang, B. Wang, J. An, M. Zhang, and Z. Zhang, "Voxel-FPN: Multiscale voxel feature aggregation for 3D object detection from LiDAR point clouds," *Sensors*, vol. 20, no. 3, p. 704, Jan. 2020.

[38] A. Newell, Z. Huang, and J. Deng, "Associative embedding: End-to-end learning for joint detection and grouping," in *Proc. Adv. Neural Inf. Process. Syst.*, Jan. 2017, pp. 2277–2287.

[39] X. Zhou, A. Karpur, L. Luo, and Q. Huang, "StarMap for category-agnostic keypoint and viewpoint estimation," in *Proc. Eur. Conf. Comput. Vis.*, 2018, pp. 318–334.

[40] B. Wu, A. Wan, X. Yue, and K. Keutzer, "SqueezeSeg: Convolutional neural nets with recurrent CRF for real-time road-object segmentation from 3D LiDAR point cloud," in *Proc. IEEE Int. Conf. Robot. Automat. (ICRA)*, May 2018, pp. 1887–1893.

[41] C. R. Qi, W. Liu, C. Wu, H. Su, and L. J. Guibas, "Frustum pointnets for 3D object detection from RGB-D data," in *Proc. IEEE Conf. Comput. Vis. Pattern Recognit.*, Jun. 2018, pp. 918–927.

[42] Z. Wang and K. Jia, "Frustum ConvNet: Sliding frustums to aggregate local point-wise features for amodal 3D object detection," 2019, *arXiv:1903.01864*. [Online]. Available: http://arxiv.org/abs/1903.01864

[43] S. Shi, Z. Wang, J. Shi, X. Wang, and H. Li, "From points to parts: 3D object detection from point cloud with part-aware and part-aggregation network," *IEEE Trans. Pattern Anal. Mach. Intell.*, vol. 43, no. 8, pp. 2647–2664, Aug. 2021.

[44] W. Wang, R. Yu, Q. Huang, and U. Neumann, "SGPN: Similarity group proposal network for 3D point cloud instance segmentation," in *Proc. IEEE/CVF Conf. Comput. Vis. Pattern Recognit.*, Jun. 2018, pp. 2569–2578.

[45] B. Yang *et al.*, "Learning object bounding boxes for 3D instance segmentation on point clouds," in *Proc. Adv. Neural Inf. Process. Syst.*, Jun. 2019, pp. 6740–6749.

[46] C. R. Qi, O. Litany, K. He, and L. J. Guibas, "Deep Hough voting for 3D object detection in point clouds," in *Proc. IEEE Int. Conf. Comput. Vis.*, Apr. 2019, pp. 9277–9286.

[47] C. R. Qi, L. Yi, H. Su, and L. J. Guibas, "PointNet++: Deep hierarchical feature learning on point sets in a metric space," in *Proc. Adv. Neural Inf. Process. Syst.*, vol. 31, Jun. 2018, pp. 5099–5108.

[48] S. Ren, K. He, R. Girshick, and J. Sun, "Faster R-CNN: Towards real-time object detection with region proposal networks," in *Proc. Adv. Neural Inf. Process. Syst.*, 2015, pp. 91–99.

[49] A. Geiger, P. Lenz, C. Stiller, and R. Urtasun, "Vision meets robotics: The KITTI dataset," *Int. J. Robot. Res.*, vol. 32, no. 11, pp. 1231–1237, 2013.

[50] H. Caesar *et al.*, "NuScenes: A multimodal dataset for autonomous driving," 2019, *arXiv:1903.11027*. [Online]. Available: http://arxiv.org/abs/1903.11027

[51] X. Chen, H. Ma, J. Wan, B. Li, and T. Xia, "Multi-view 3D object detection network for autonomous driving," in *Proc. IEEE Conf. Comput. Vis. Pattern Recognit.*, Jul. 2017, pp. 1907–1915.

[52] J. Ku, M. Mozifian, J. Lee, A. Harakeh, and S. L. Waslander, "Joint 3D proposal generation and object detection from view aggregation," in *Proc. IEEE/RSJ Int. Conf. Intell. Robots Syst.*, Oct. 2018, pp. 1–8.

[53] M. Liang, B. Yang, Y. Chen, R. Hu, and R. Urtasun, "Multi-task multi-sensor fusion for 3D object detection," in *Proc. IEEE/CVF Conf. Comput. Vis. Pattern Recognit. (CVPR)*, Jun. 2019, pp. 7345–7353.

[54] Z. Liu, X. Zhao, T. Huang, R. Hu, Y. Zhou, and X. Bai, "TANet: Robust 3D object detection from point clouds with triple attention," in *Proc. AAAI Conf. Artif. Intell.*, 2020, pp. 11677–11684.

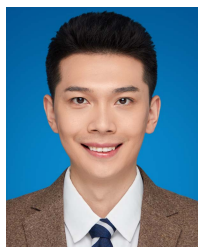

**Siyu Teng** received the master's degree from Jilin University in 2021. He is currently pursuing the Ph.D. degree with HKBU.

His current research interests include path planning, imitation learning, and object detection.

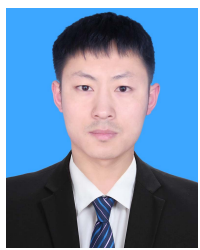

**Guojun Wang** received the B.S. degree in vehicle engineering from Yanshan University, Qinhuangdao, China, in 2014. He is currently pursuing the Ph.D. degree in vehicle engineering with Jilin University, Changchun, China.

His current research interests include computer vision, deep learning, and autonomous driving.

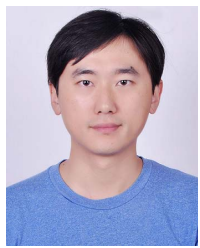

**Long Chen** (Senior Member, IEEE) received the B.S. degree in communication engineering and the Ph.D. degree in signal and information processing from Wuhan University, Wuhan, China. He is currently an Associate Professor with the School of Data and Computer Science, Sun Yat-sen University, Guangzhou, China. His research interests include autonomous driving, robotics, and artificial intelligence, where he has contributed more than 70 publications. He received the IEEE Vehicular Technology Society 2018 Best Land Transportation Paper Award. He also serves as an Associate Editor for IEEE TRANSACTIONS ON INTELLIGENT TRANSPORTATION SYSTEMS.

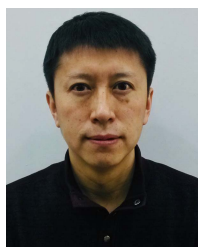

**Jian Wu** is currently a Professor with the College of Automotive Engineering, Jilin University. He is the author of over 40 peers-reviewed papers in international journals and conferences. He has been in charge of numerous projects funded by national government and institutional organizations on vehicles. His research interests include vehicle control systems, electric vehicles, and intelligent vehicles.

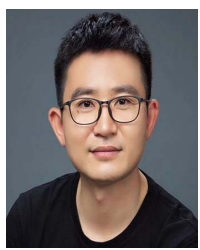

**Dongpu Cao** (Senior Member, IEEE) received the Ph.D. degree from Concordia University, Montreal, QC, Canada, in 2008. He is currently the Canada Research Chair of driver cognition and automated driving. He is also an Associate Professor and the Director of Waterloo Cognitive Autonomous Driving (CogDrive) Lab, University of Waterloo, Waterloo, ON, Canada. He has contributed more than 200 articles and three books. His current research interests include driver cognition, automated driving, and cognitive autonomous driving.

He received the SAE Arch T. Colwell Merit Award in 2012, the IEEE VTS 2020 Best Vehicular Electronics Paper Award, and three Best Paper Awards from ASME and IEEE conferences. He also serves on the SAE Vehicle Dynamics Standards Committee and acts as the Co-Chair for IEEE ITSS Technical Committee on Cooperative Driving. He serves as the Deputy Editor-in-Chief for *IET Intelligent Transport Systems* journal and an Associate Editor for IEEE TRANSACTIONS ON VEHICULAR TECHNOLOGY, IEEE TRANSACTIONS ON INTELLIGENT TRANSPORTATION SYSTEMS, IEEE/ASME TRANSACTIONS ON MECHATRONICS, IEEE TRANSACTIONS ON INDUSTRIAL ELECTRONICS, IEEE/CAA JOURNAL OF AUTOMATICA SINICA, IEEE TRANSACTIONS ON COMPUTATIONAL SOCIAL SYSTEMS, and *ASME Journal of Dynamic Systems, Measurement and Control*. He was a Guest Editor of *Vehicle System Dynamics*, IEEE TRANSACTIONS ON SYSTEMS, MAN, AND CYBERNETICS: SYSTEMS, and IEEE INTERNET OF THINGS JOURNAL. He is also an IEEE VTS Distinguished Lecturer.

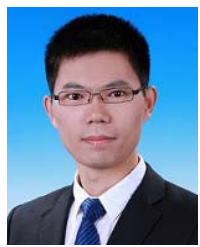

**Bin Tian** received the B.S. degree from Shandong University, Jinan, China, in 2009, and the Ph.D. degree from the Institute of Automation, Chinese Academy of Sciences, Beijing, China, in 2014.

He is currently an Associate Professor with the State Key Laboratory of Management and Control for Complex Systems, Institute of Automation, Chinese Academy of Sciences. His current research interests include computer vision, machine learning, and automated driving.